\documentclass[runningheads]{llncs}
\usepackage[T1]{fontenc}
\usepackage[misc]{ifsym}
\usepackage{graphicx}
\usepackage{amsmath}
\usepackage{multirow}
\usepackage{xcolor}
\usepackage{amssymb}
\usepackage{algorithm}
\usepackage{algorithmic}
\usepackage{pifont}

\begin{document}
\title{A Large Language Model Guided Topic Refinement Mechanism for Short Text Modeling}
%
%\titlerunning{Abbreviated paper title}
% If the paper title is too long for the running head, you can set
% an abbreviated paper title here
%

\author{Shuyu Chang\inst{1}\textsuperscript{,*} \and
Rui Wang\inst{1}\textsuperscript{,*} \and
Peng Ren\inst{1} \and
Qi Wang\inst{1} \and
Haiping Huang\inst{1,2}\textsuperscript{(\Letter)}}

\authorrunning{Chang et al.}
% First names are abbreviated in the running head.
% If there are more than two authors, 'et al.' is used.
%
\institute{Nanjing University of Posts and Telecommunications, Nanjing, Jiangsu, China \and
Jiangsu High Technology Research Key Laboratory for Wireless Sensor Networks, Nanjing, Jiangsu, China\\
\email{\{shuyu\_chang,rui\_wang,1022041122,1223045440,hhp\}@njupt.edu.cn}}

\titlerunning{A LLM-Guided Topic Refinement Mechanism for Short Text Modeling}
\maketitle              % typeset the header of the contribution

{\renewcommand*{\thefootnote}{}
   \footnotetext{\textsuperscript{*} These authors contributed equally to this work.}}

\begin{abstract}
Modeling topics effectively in short texts, such as tweets and news snippets, is crucial to capturing rapidly evolving social trends.
Existing topic models often struggle to accurately capture the underlying semantic patterns of short texts, primarily due to the sparse nature of such data.
This nature of texts leads to an unavoidable lack of co-occurrence information, which hinders the coherence and granularity of mined topics.
This paper introduces a novel model-agnostic mechanism, termed Topic Refinement, which leverages the advanced text comprehension capabilities of Large Language Models (LLMs) for short-text topic modeling.
Unlike traditional methods, this post-processing mechanism enhances the quality of topics extracted by various topic modeling methods through prompt engineering.
We guide LLMs in identifying semantically intruder words within the extracted topics and suggesting coherent alternatives to replace these words.
This process mimics human-like identification, evaluation, and refinement of the extracted topics.
Extensive experiments on four diverse datasets demonstrate that Topic Refinement boosts the topic quality and improves the performance in topic-related text classification tasks.

\end{abstract}

\section{Introduction}
Short texts, such as online comments and news headlines, are influential in shaping public opinion and mirroring social trends~\cite{pang2021,zhang2022b,zhang2023,wang2024}. 
Analyzing these texts can uncover underlying topics, providing insights into social discourse. 
Topic modeling emerges as a valuable tool to extract topics from vast data~\cite{lin2020,wang2023,wu2024,wang2025}. 
Nevertheless, the sparsity of context in short texts poses challenges for conventional topic models~\cite{blei2003,yang2021}. 
This brevity leads to low information density~\cite{qiang2022}, making it difficult to capture and represent the semantics of discussed topics.

Recent advancements have alleviated the issue of data sparsity by enhancing word co-occurrence patterns~\cite{yan2013,cheng2014} and incorporating external knowledge sources~\cite{pennington2014,devlin2019} to strengthen semantic relationships. 
Furthermore, approaches like quantified distribution analysis~\cite{oord2017} and contrastive learning techniques~\cite{he2020} have been employed to improve the learning signals derived from short texts. 
Despite these efforts, existing methods frequently yield suboptimal semantic coherence and granularity (i.e., clarity and diversity from a user perspective)~\cite{mu2024} of topics in practical applications.

\begin{figure}[t]
    \centering
    \includegraphics[width=0.75\linewidth]{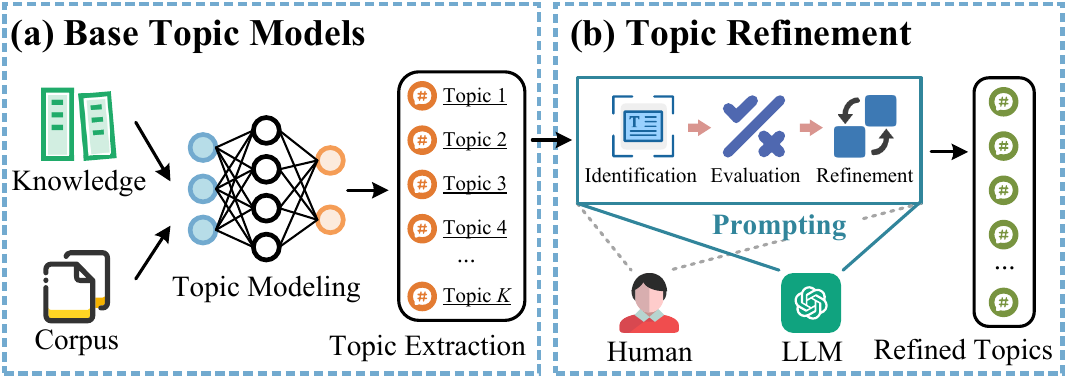}
    \caption{Workflow of base topic models and our Topic Refinement mechanism.}
    \label{fig:workflow}
\end{figure}

In light of the limitations of previous work, we shift to the booming and widely concerned Large Language Models (LLMs) such as PaLM~\cite{chowdhery2023}, GPT~\cite{brown2020,ouyang2022}, and others. 
With their unprecedented semantic comprehension capabilities~\cite{dijk2023,tan2023,ye2023}, trained across extensive datasets, LLMs are expected to further improve the quality of topic modeling for short texts. 
Existing LLM-based topic modeling frameworks like PromptTopic~\cite{wang2023b} and TopicGPT~\cite{pham2024} generate topics by inputting documents to produce topic categorizations or descriptions. 
However, they deviate from the classic bag-of-words format, complicating direct evaluation and comparisons with traditional topic modeling outputs. 
These models also require whole documents as inputs, leading to high token consumption and increasing computational costs.

To bridge this gap, we introduce a novel model-agnostic mechanism termed Topic Refinement. 
Figure~\ref{fig:workflow} illustrates the workflow of our proposed mechanism. 
In the traditional setup, topics are mined directly from the corpus with external knowledge using existing methods (called base topic models). 
Conversely, Topic Refinement does not involve direct topic mining but instead leverages LLMs via prompt engineering to refine topics initially mined by base topic models.
More concretely,  this mechanism iterates over each topic and sequentially assumes a word as the potential intruder word in a topic while the remaining words represent this topic. 
Prompting LLMs are then employed to identify the topic and assess if the word aligns with the semantic expression of the other words. 
If alignment is confirmed, the word is retained; otherwise, coherent words are generated as candidates to replace the intruder word. 
Topic Refinement mimics the human process of identifying, evaluating, and refining extracted topics to make them more coherent and accessible for users to discern.

The main contributions of our work are summarized as follows:
\begin{itemize}
    \item We introduce Topic Refinement, a model-agnostic mechanism for short text topic modeling, pioneering a new pathway to improve topic quality after initial mining.
    \item Topic Refinement innovatively integrates LLMs with diverse base topic models to precisely evaluate and refine the representative words for each extracted topic.
    \item Extensive experiments are conducted on four datasets to evaluate the effectiveness of our mechanism. Our results prove improvements over existing methods across various metrics.
\end{itemize}

The remainder of this paper is organized as follows.
Section~\ref{sec:related} reviews some related works.
We describe our proposed mechanism in Section~\ref{sec:method}.
Experiments and analysis are presented in Section~\ref{sec:experiments}, and we conclude our work in Section \ref{sec:conclusion}.

\section{Related work}
\label{sec:related}
This paper intersects with two primary research areas: short text topic modeling and LLMs.

\subsection{Short Text Topic Modeling}
Researchers initially relied on methods such as LDA~\cite{blei2003} but encountered challenges in applying them to short texts.
This led to the development of BTM~\cite{cheng2014}, which addressed the limitations by exploring broader linguistic patterns by expanding co-occurring word pairs.
Researchers began incorporating word embeddings into topic modeling as the field progressed to capture semantic relationships more effectively~\cite{li2016,wu2023,zuo2023}.
Approaches like BERTopic~\cite{grootendorst2022} and TopClus~\cite{meng2022} further utilized pre-trained language models to extract contextualized embeddings for topic modeling~\cite{fang2024}.
Furthermore, techniques involving labels~\cite{churchill2022} and document connectivity~\cite{zhang2022a} have provided structure-based approaches to improve topic quality.
Another promising direction is the generation of quantified topic distributions tailored for short texts~\cite{wu2020,wu2022}. 
These methods aggregate short texts with similar topics into the same topic distribution.
Recently, contrastive learning has been explored to further mitigate the effects of data sparsity~\cite{nguyen2021,han2023,zhou2023}. 
These models rely on data augmentation to learn from similar and dissimilar text pairs, enhancing the learning signals from short texts.
Despite these advancements, existing approaches often fail to achieve satisfactory topic quality.
To address this, our study introduces a model-agnostic refinement mechanism that enhances the results of base topic models via LLMs.

\begin{figure*}[ht]
  \centering
  \includegraphics[width=0.99\linewidth]{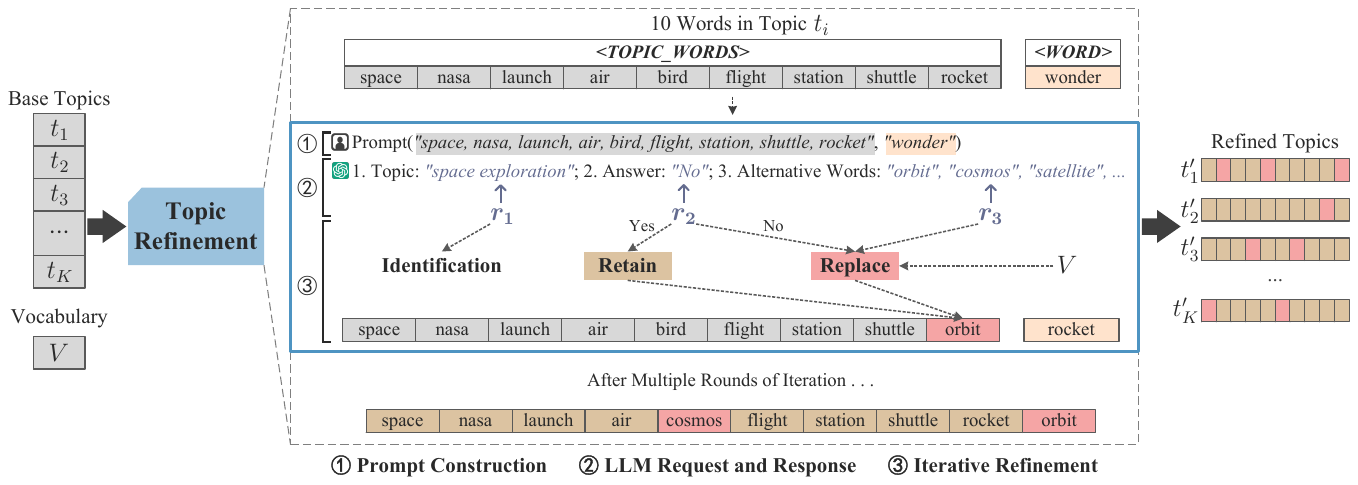}
  \caption{Overview of Topic Refinement ($N=10$): \ding{172} Prompt Construction; \ding{173} LLM Request and Response; \ding{174} Iterative Refinement.}
  \label{fig:overview}
\end{figure*}

\subsection{Large Language Models}
LLMs have increasingly been explored for their robust capabilities in various natural language processing tasks~\cite{zhang2023b}. 
Studies~\cite{chowdhery2023,brown2020,ouyang2022} have demonstrated that LLMs excel at contextual understanding and generating human-like text.
Their ability to grasp subtle semantic nuances offers potential improvements in topic modeling for short texts. 
Recently, there has been some work that utilizes LLMs for topic modeling~\cite{mu2024,wan2024}. 
For instance, the PromptTopic~\cite{wang2023b} and TopicGPT~\cite{pham2024} frameworks extract topics from individual documents by generating topic categorizations and descriptions. 
However, their output is represented by phrases or sentences, which differs from traditional topic modeling, which uses a bag of words. 
This difference in output representation poses challenges in directly comparing the results with those of traditional models.
Additionally, these frameworks require feeding entire documents as inputs, which can consume many tokens.
Our approach adheres to the bag-of-words format to represent topics and harnesses LLMs to optimize topic quality through efficient refinement.

\section{Methodology}
\label{sec:method}

\subsection{Problem Definition}
Topic Refinement aims to enhance the coherence and granularity of topics generated from base topic models for short texts.
Formally, given a set of topics $T=\left\{t_1, t_2, \cdots, t_K \right\}$ discovered by a base topic model, each topic consists of $N$ representative words, denoted by $t_i=\left\{w_{i1}, w_{i2}, \cdots w_{iN}\right\}$, where $1\leq i\leq K$.
In this context, an intruder word is defined as a word within a topic that is not semantically coherent with the other words, thus disrupting the quality of this topic.
The Topic Refinement mechanism seeks to transform each base topic $t_i$ into a refined topic $t'_i$ by correcting such intruder words $w_{ij}$ in its word set $t_i$.
This transformation is guided by a refinement function $\mathcal{R}$, which determines whether to retain or replace the potentially intruder word $w_{ij}$ in $t_i$ as follows:
\begin{equation}
    \mathcal{R}: \left(t_i\setminus\{w_{ij}\}, w_{ij}\right) \to w'_{ij},
\end{equation}
where $w'_{ij}$ is the alternative word for topic $t_i$.
We propose to model the refinement function via LLMs, leveraging their advanced capabilities in text understanding.

\subsection{Topic Refinement}
Figure~\ref{fig:overview} depicts the overall architecture of our Topic Refinement mechanism.
Before this mechanism, we extract a set of topics $T=\left\{t_1, t_2, \cdots, t_K \right\}$ and the vocabulary $V$ from a short text dataset through a base topic model.
This base model can utilize existing methods in this field, such as probabilistic topic models or neural topic models.
Consequently, our mechanism unfolds in three phases: prompt construction, LLM request and response, and iterative refinement.

\begin{figure}[t]
  \centering
  \includegraphics[width=0.95\linewidth]{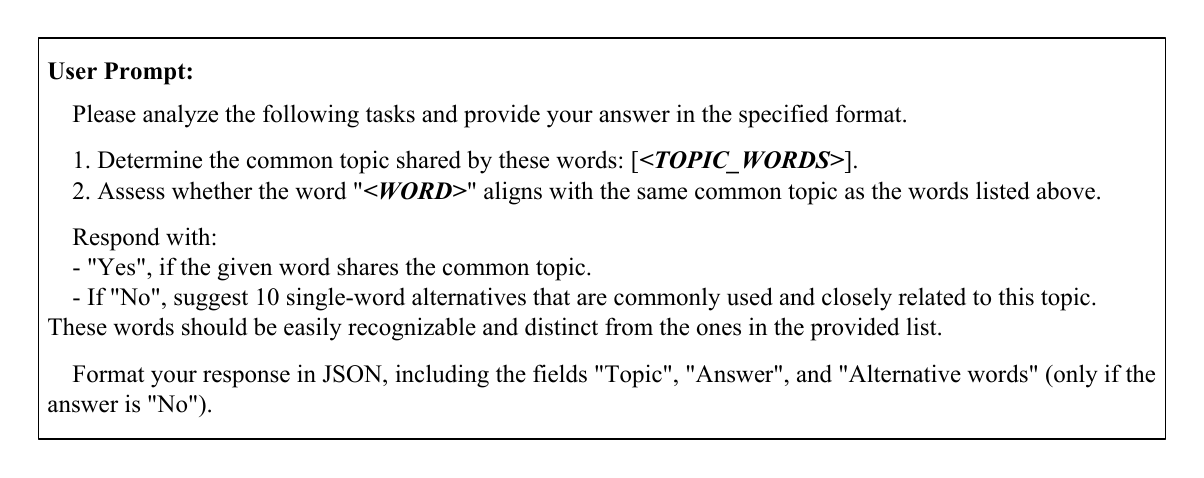}
  \caption{LLM prompt template for Topic Refinement.}
  \label{fig:prompt}
\end{figure}

\subsubsection{Prompt Construction}
Prompt engineering serves as the direct way to guide and control the output generated by LLMs.
To model the refinement function $\mathcal{R}$, our prompt template is designed to guide LLMs to find intruder words in each topic and generate alternative words related to this topic. 
Given the topic $t_i$ and the words $\left\{w_{i1}, w_{i2}, \cdots w_{iN}\right\}$ in it, we structure the prompt template into two tasks.
Firstly, we employ LLM $\mathcal{M}$ to identify the common topic shared by the \textit{<TOPIC\_WORDS>}, which is represented by words $t_i\setminus\{w_{ij}\}$. 
Later, LLM $\mathcal{M}$ assesses whether the word $w_{ij}$, denoted as \textit{<WORD>}, is semantically consistent with the previously determined common topic.
In cases of inconsistency, LLM $\mathcal{M}$ is also tasked with suggesting ten commonly used words closely related to that topic as candidate alternatives. 
For the clarity and diversity of topics, these candidate words should be easily recognizable and distinctly different from those in the provided list.
Moreover, to facilitate subsequent processing of generated results, the prompt template also enforces the format of generated texts to be in JSON. 
The prompt template is defined as $p = \mathrm{Prompt}$(\textit{<TOPIC\_WORDS>}, \textit{<WORD>}), which is used to construct each iteration of prompts $p$ for LLM $\mathcal{M}$.
The details of the prompt template are presented in Figure~\ref{fig:prompt}.

\subsubsection{LLM Request and Response}
After constructing the prompt $p=\mathrm{Prompt}(t_i\setminus\{w_{ij}\}, w_{ij})$, we initiate requests to LLM $\mathcal{M}$ for its response. 
Essentially, the LLM models the Topic Refinement function by estimating the conditional probability of the generated text: $\mathrm{P_{LLM}}(s_1, s_2, \cdots, s_n  \mid p)$, where $(s_1, s_2, \cdots, s_n)$ is the generated word sequence with variable lengths.
The generated sequence, formatted in JSON, is then converted into our desired results by a simple mapping function $(r_1, r_2, r_3)=f(s_1, s_2, \cdots, s_n)$.
$r_1$ represents the topic identified by LLM $\mathcal{M}$, typically composed of one word or phrase summarizing this topic. 
$r_2$ is the judgment of whether the \textit{<WORD>} in the prompt $p$ is semantically coherent with the identified topic, with the response being either "\textit{Yes}" or "\textit{No}". 
$r_3$ is a list of ten candidate words for replacement, generated only if $r_2$ is "\textit{No}"; otherwise, it remains empty.
The process above is formalized as follows:
\begin{equation}
\label{eq:llm}
    (r_1, r_2, r_3) = \mathcal{M}(\mathrm{Prompt}(t_i\setminus\{w_{ij}\}, w_{ij})).
\end{equation}

\begin{algorithm}[t]
\caption{Topic Refinement Mechanism}
\label{alg:tr}
\begin{algorithmic}
\REQUIRE Topics $T$, Vocabulary $V$, LLM $\mathcal{M}$
\ENSURE Refined topics $T'=\left\{t'_1, t'_2, \cdots, t'_K \right\}$
\STATE Initialize the set of refined topics $T' \gets \varnothing$;
\FOR{each topic $t_i$ in $T$}
    \STATE Initialize the refined topic $t'_i \gets t_i$;
    \FOR{each word $(w_{iN}, w_{i(N-1)}, \cdots, w_{i1})$ in $t'_i$}
        \STATE Construct the prompt $p \gets\mathrm{Prompt}(t_i\setminus\{w_{ij}\}, w_{ij})$;
        \STATE Generate the answer $(r_1, r_2, r_3)$ by Eq.~\ref{eq:llm};
        \IF{the answer $r_2$ is "\textit{NO}"}
            \STATE Select the alternative word $w'_{ij}$ from $r_3$ and $V$;
            \STATE Replace $w_{ij}$ in topic $t'_i$ with $w'_{ij}$;
        \ELSE
            \STATE Retain $w_{ij}$ in topic $t'_i$;
    \ENDIF
    \ENDFOR
\ENDFOR
\end{algorithmic}
\end{algorithm}

\subsubsection{Iterative Refinement}
This phase, informed by the results from LLM $\mathcal{M}$, selects a more coherent alternative to replace the intruder word when necessary.
If $r_2$ indicates that a word $w_{ij}$ lies outside the semantic boundary of the topic $t_i$, a list of ten alternative words is generated in $r_3$. 
The alternative word $w'_{ij}$ is selected from this list based on its inclusion in the vocabulary $V$.
Assume that none of the alternative words are found in the vocabulary.
In that case, the mechanism selects the word from $V$ that is most semantically similar on average to the alternative words and is not already included in the topic $t_i$.
This ensures that the alternative word $w'_{ij}$ is not only semantically coherent but also retains the lexical integrity of the dataset.
The process iterates over each word in a topic in reverse order,  for words are generally arranged according to topic relevance.
By the end of this iteration, we arrive at a refined set of topics $t'_i$.
The specific procedure of the Topic Refinement mechanism is demonstrated in Algorithm \ref{alg:tr}.

\section{Experiments}
\label{sec:experiments}
In this section, we conduct extensive experiments to compare the performance of different base models before and after refinement. 
We also show text classification, ablation study, and visualization to demonstrate the effectiveness of Topic Refinement.

\subsection{Experimental Setup}
\paragraph{Datasets.} We employ the following four short text datasets for evaluation:
(1) \textbf{Tweet}~\cite{yin2016}: tweets that are highly relevant to 89 queries, manually labeled in the 2011 and 2012 microblog tracks at the Text REtrieval Conference.
(2) \textbf{AGNews}~\cite{zhang2015}: more than one million news articles across four categories gathered by ComeToMyHead. We randomly sample 5,000 article titles from each category for experiments.
(3) \textbf{TagMyNews}~\cite{vitale2012}: news titles from various news sources, annotated with seven labels.
(4) \textbf{YahooAnswer}~\cite{zhang2015}: ten categories of questions and their corresponding best answers from the Yahoo Answers website. Similar to AGNews, we sample 2,500 question titles from each category.
Table~\ref{tab:datasets} shows the statistics of these datasets after preprocessing with reference to Wu~\textit{et al.}\cite{wu2022}.

\begin{table}[t]
    \centering
    \caption{Statistics of datasets after preprocessing.}
    \label{tab:datasets}
    \resizebox{0.55\linewidth}{!}{
    \begin{tabular}{|l|c|c|c|c|}
    \hline
    \textbf{Datasets} & \textbf{\#Docs} & \textbf{\#Labels} & \textbf{\#Vocab.} & \textbf{Avg. Length} \\ 
    \hline
    \textit{Tweet}        & 2,133     & 89  & 1,127   & 5.550 \\
    \textit{AGNews}       & 14,845    & 4   & 3,290   & 4.268 \\
    \textit{TagMyNews}    & 27,369    & 7   & 4,325   & 4.483 \\
    \textit{YahooAnswer}  & 12,258    & 10  & 3,423   & 4.151 \\
    \hline
\end{tabular}}
\end{table}

\paragraph{Base Models.}
We utilize the Topic Refinement with the following base topic models:
(1) \textbf{LDA}~\cite{blei2003} that uses Gibbs sampling for parameter estimation and inference, with GibbsLDA++\footnote{\url{https://gibbslda.sourceforge.net}} implementation in our experiment. 
(2) \textbf{BTM}~\cite{cheng2014} that is designed for short texts to enrich word co-occurrence within them\footnote{\url{https://github.com/xiaohuiyan/BTM}}.
(3) \textbf{Gaussian-BAT}~\cite{wang2020} that incorporates word relatedness information by multivariate Gaussian modeling of topics.
(4) \textbf{CNTM}~\cite{nguyen2021} that leverages contrastive learning and word embedding to capture semantic patterns of documents\footnote{\url{https://github.com/nguyentthong/CLNTM}}.
(5) \textbf{TSCTM}~\cite{wu2022} that is based on contrastive learning with quantified sampling strategies to improve topic quality in short texts\footnote{\url{https://github.com/bobxwu/TSCTM}}.
(6) \textbf{BERTopic}~\cite{grootendorst2022} that utilizes pre-trained transformer-based language models to extract topics\footnote{\url{https://maartengr.github.io/BERTopic}}.
(7) \textbf{ECRTM}~\cite{wu2023} that forces each topic embedding to be the center of a word embedding cluster in the semantic space via optimal transport\footnote{\url{https://github.com/BobXWu/ECRTM}}. 
(8) \textbf{CWTM}~\cite{fang2024} that integrates contextualized word embeddings from BERT to learn the topic vector of a document without BOW information\footnote{\url{https://github.com/Fitz-like-coding/CWTM}}.
(9) \textbf{LLM-TM}~\cite{mu2024} that prompts LLMs to generate topic titles from a given set of documents. 
We adapt the original prompts to generate a bag of words for each topic, allowing for a standardized evaluation of topic quality\footnote{\url{https://github.com/GateNLP/LLMs-for-Topic-Modeling}}.
All base models are executed with their recommended hyperparameters.

\paragraph{Evaluation Metrics.}
Following previous mainstream work, we evaluate the topic quality by topic coherence metrics, measuring the coherence between the $N$ words of extracted topics.
We adopt the widely-used Palmetto\footnote{\url{https://github.com/dice-group/Palmetto}} to calculate these coherence metrics $C_A$, $C_P$, $C_V$, UCI, and NPMI~\cite{roder2015}, with higher values indicating better topic quality. 
Inspired by Mu \textit{et al.}~\cite{mu2024}, we also introduce two topic granularity metrics based on word embeddings to evaluate topic clarity and diversity: within-topic similarity $\mathcal{S}$ and between-topic distance $\mathcal{D}$. 
These metrics evaluate the semantic similarity of words within individual topics and the distinction between different topics, respectively. 
The calculations are as follows:
\begin{equation}
    \mathcal{S} = \frac{2}{KN(N-1)} \!\sum_{i}^{K} \sum_{j}^{N-1} \!\sum_{k=j+1}^{N} \frac{e_{w_{ij}} \cdot e_{w_{ik}}}{\left\| e_{w_{ij}}\right\|_2 \left\| e_{w_{ik}}\right\|_2 },
\end{equation}
\begin{equation}
    \mathcal{D} = \frac{2}{K(K-1)} \sum_{i}^{K-1} \sum_{m=i+1}^{K} \sum_{d}^{D} \left | e_{t_i}^{(d)} - e_{t_m}^{(d)} \right |^2,
\end{equation}
where $e_{w_{ij}}$ is the embedding of word $w_{ij}$, and $D$ is the dimension of word embeddings. $e_{t_i}=\frac{1}{N}\sum_{j}^{N} e_{w_{ij}}$ is the centroid of topic $t_i$.
The symbol $\Delta$ represents the changes in metric after refinement.
To evaluate Topic Refinement in downstream tasks, we also employ the topic distribution of each document for text classification. 
The experiments evaluate classification performance based on Accuracy and F1 score metrics.

\paragraph{Implementation Details.}
For our Topic Refinement mechanism, we use GPT-3.5-turbo as the LLM $\mathcal{M}$ and set its temperature as 0 to lower the completion randomness.
Keeping with prior work, each topic $t_i$ mined by base topic models is represented by $N$ words, where $N$ is typically set to 10.
The evaluation metrics based on word embeddings are calculated utilizing the GloVe.840B.300d\footnote{\url{https://nlp.stanford.edu/data/glove.840B.300d.zip}}, where $D = 300$.

\begin{table*}[t]
    \centering
    \caption{Topic coherence results on four datasets.}
    \label{tab:main}
\resizebox{0.9\linewidth}{!}{
\begin{tabular}{|l|l|cc|cc|cc|cc|cc|}
\hline
\textbf{Datasets}                     & \textbf{Base Models} & $\boldsymbol{C_A}$ & $\boldsymbol{\Delta_{C_A}}$ & $\boldsymbol{C_P}$ & $\boldsymbol{\Delta_{C_P}}$ & $\boldsymbol{C_V}$ & $\boldsymbol{\Delta_{C_V}}$ & $\textbf{UCI}$ & $\boldsymbol{\Delta_{\text{\textbf{UCI}}}}$ & $\textbf{NPMI}$ & $\boldsymbol{\Delta_{\text{\textbf{NPMI}}}}$ \\ \hline
\multirow{9}{*}{\textit{Tweet}}       & LDA       & 0.175   & {\color{blue} \textbf{+}}0.045   & 0.142   & {\color{blue} \textbf{+}}0.223   & 0.375   & {\color{blue} \textbf{+}}0.022   & -0.675  & {\color{blue} \textbf{+}}1.452   & 0.001   & {\color{blue} \textbf{+}}0.086   \\
                                      & BTM       & 0.191   & {\color{blue} \textbf{+}}0.022   & 0.152   & {\color{blue} \textbf{+}}0.122   & 0.377   & {\color{blue} \textbf{+}}0.017   & -0.682  & {\color{blue} \textbf{+}}0.705   & 0.002   & {\color{blue} \textbf{+}}0.040   \\
                                      & G-BAT     & 0.160   & {\color{blue} \textbf{+}}0.035   & -0.005  & {\color{blue} \textbf{+}}0.187   & 0.419   & {\color{red}  \textbf{--}}0.012  & -1.992  & {\color{blue} \textbf{+}}1.348   & -0.052  & {\color{blue} \textbf{+}}0.065   \\
                                      & CNTM      & 0.183   & {\color{blue} \textbf{+}}0.032   & 0.153   & {\color{blue} \textbf{+}}0.120   & 0.382   & {\color{blue} \textbf{+}}0.016   & -0.744  & {\color{blue} \textbf{+}}0.675   & 0.001   & {\color{blue} \textbf{+}}0.040   \\
                                      & TSCTM     & 0.182   & {\color{blue} \textbf{+}}0.027   & 0.134   & {\color{blue} \textbf{+}}0.158   & 0.392   & {\color{blue} \textbf{+}}0.014   & -0.993  & {\color{blue} \textbf{+}}0.900   & -0.008  & {\color{blue} \textbf{+}}0.050   \\
                                      & BERTopic  & 0.205   & {\color{blue} \textbf{+}}0.016   & 0.203   & {\color{blue} \textbf{+}}0.089   & 0.394   & {\color{blue} \textbf{+}}0.011   & -0.644  & {\color{blue} \textbf{+}}0.588   & 0.012   & {\color{blue} \textbf{+}}0.032   \\
                                      & ECRTM     & 0.191   & {\color{blue} \textbf{+}}0.026   & 0.109   & {\color{blue} \textbf{+}}0.152   & 0.421   & {\color{blue} \textbf{+}}0.002   & -1.920  & {\color{blue} \textbf{+}}1.182   & -0.039  & {\color{blue} \textbf{+}}0.058   \\
                                      & CWTM      & 0.211   & {\color{blue} \textbf{+}}0.021   & 0.191   & {\color{blue} \textbf{+}}0.157   & 0.415   & {\color{blue} \textbf{+}}0.012   & -1.342  & {\color{blue} \textbf{+}}1.322   & -0.013  & {\color{blue} \textbf{+}}0.065   \\
                                      & LLM-TM    & 0.154   & {\color{blue} \textbf{+}}0.035   & -0.026  & {\color{blue} \textbf{+}}0.206   & 0.392   & {\color{red}  \textbf{--}}0.005  & -1.964  & {\color{blue} \textbf{+}}1.443   & -0.059  & {\color{blue} \textbf{+}}0.072   \\ \hline
\multirow{9}{*}{\textit{AGNews}}      & LDA       & 0.174   & {\color{blue} \textbf{+}}0.026   & 0.220   & {\color{blue} \textbf{+}}0.110   & 0.373   & {\color{blue} \textbf{+}}0.021   & 0.062   & {\color{blue} \textbf{+}}0.410   & 0.036   & {\color{blue} \textbf{+}}0.028   \\
                                      & BTM       & 0.166   & {\color{blue} \textbf{+}}0.039   & 0.153   & {\color{blue} \textbf{+}}0.196   & 0.367   & {\color{blue} \textbf{+}}0.028   & -0.034  & {\color{blue} \textbf{+}}0.736   & 0.023   & {\color{blue} \textbf{+}}0.051   \\
                                      & G-BAT     & 0.181   & {\color{blue} \textbf{+}}0.023   & 0.192   & {\color{blue} \textbf{+}}0.145   & 0.398   & {\color{blue} \textbf{+}}0.005   & -0.622  & {\color{blue} \textbf{+}}0.837   & 0.012   & {\color{blue} \textbf{+}}0.044   \\
                                      & CNTM      & 0.266   & {\color{blue} \textbf{+}}0.013   & 0.469   & {\color{blue} \textbf{+}}0.059   & 0.448   & {\color{blue} \textbf{+}}0.010   & 0.744   & {\color{blue} \textbf{+}}0.275   & 0.101   & {\color{blue} \textbf{+}}0.017   \\
                                      & TSCTM     & 0.205   & {\color{blue} \textbf{+}}0.033   & 0.155   & {\color{blue} \textbf{+}}0.224   & 0.437   & {\color{blue} \textbf{+}}0.005   & -1.705  & {\color{blue} \textbf{+}}1.704   & -0.022  & {\color{blue} \textbf{+}}0.082   \\
                                      & BERTopic  & 0.228   & {\color{blue} \textbf{+}}0.029   & 0.295   & {\color{blue} \textbf{+}}0.138   & 0.442   & {\color{blue} \textbf{+}}0.009   & -0.458  & {\color{blue} \textbf{+}}0.859   & 0.038   & {\color{blue} \textbf{+}}0.045   \\
                                      & ECRTM     & 0.203   & {\color{blue} \textbf{+}}0.051   & 0.125   & {\color{blue} \textbf{+}}0.339   & 0.443   & {\color{blue} \textbf{+}}0.004   & -2.387  & {\color{blue} \textbf{+}}2.630   & -0.053  & {\color{blue} \textbf{+}}0.125   \\
                                      & CWTM      & 0.173   & {\color{blue} \textbf{+}}0.017   & 0.206   & {\color{blue} \textbf{+}}0.094   & 0.387   & {\color{blue} \textbf{+}}0.006   & -0.678  & {\color{blue} \textbf{+}}0.606   & 0.005   & {\color{blue} \textbf{+}}0.032   \\
                                      & LLM-TM    & 0.166   & {\color{blue} \textbf{+}}0.042   & 0.081   & {\color{blue} \textbf{+}}0.206   & 0.362   & {\color{blue} \textbf{+}}0.030   & -1.296  & {\color{blue} \textbf{+}}1.262   & -0.033  & {\color{blue} \textbf{+}}0.072   \\ \hline
\multirow{9}{*}{\textit{TagMyNews}}   & LDA       & 0.189   & {\color{blue} \textbf{+}}0.027   & 0.237   & {\color{blue} \textbf{+}}0.147   & 0.391   & {\color{blue} \textbf{+}}0.018   & 0.021   & {\color{blue} \textbf{+}}0.594   & 0.038   & {\color{blue} \textbf{+}}0.039   \\
                                      & BTM       & 0.177   & {\color{blue} \textbf{+}}0.039   & 0.142   & {\color{blue} \textbf{+}}0.215   & 0.389   & {\color{blue} \textbf{+}}0.020   & -0.594  & {\color{blue} \textbf{+}}1.039   & 0.005   & {\color{blue} \textbf{+}}0.061   \\
                                      & G-BAT     & 0.216   & {\color{blue} \textbf{+}}0.021   & 0.332   & {\color{blue} \textbf{+}}0.082   & 0.407   & {\color{blue} \textbf{+}}0.017   & 0.302   & {\color{blue} \textbf{+}}0.417   & 0.061   & {\color{blue} \textbf{+}}0.026   \\
                                      & CNTM      & 0.293   & {\color{blue} \textbf{+}}0.010   & 0.515   & {\color{blue} \textbf{+}}0.060   & 0.475   & {\color{blue} \textbf{+}}0.009   & 0.608   & {\color{blue} \textbf{+}}0.421   & 0.103   & {\color{blue} \textbf{+}}0.021   \\
                                      & TSCTM     & 0.209   & {\color{blue} \textbf{+}}0.033   & 0.153   & {\color{blue} \textbf{+}}0.179   & 0.465   & {\color{blue} \textbf{+}}0.003   & -1.911  & {\color{blue} \textbf{+}}1.343   & -0.027  & {\color{blue} \textbf{+}}0.066   \\
                                      & BERTopic  & 0.217   & {\color{blue} \textbf{+}}0.018   & 0.287   & {\color{blue} \textbf{+}}0.129   & 0.459   & {\color{blue} \textbf{+}}0.010   & -1.090  & {\color{blue} \textbf{+}}0.833   & 0.012   & {\color{blue} \textbf{+}}0.042   \\
                                      & ECRTM     & 0.190   & {\color{blue} \textbf{+}}0.045   & 0.086   & {\color{blue} \textbf{+}}0.261   & 0.451   & {\color{red}  \textbf{--}}0.008  & -2.623  & {\color{blue} \textbf{+}}2.114   & -0.064  & {\color{blue} \textbf{+}}0.100   \\
                                      & CWTM      & 0.198   & {\color{blue} \textbf{+}}0.016   & 0.244   & {\color{blue} \textbf{+}}0.079   & 0.407   & {\color{blue} \textbf{+}}0.009   & -0.542  & {\color{blue} \textbf{+}}0.475   & 0.019   & {\color{blue} \textbf{+}}0.027   \\
                                      & LLM-TM    & 0.158   & {\color{blue} \textbf{+}}0.046   & 0.047   & {\color{blue} \textbf{+}}0.254   & 0.382   & {\color{blue} \textbf{+}}0.024   & -1.623  & {\color{blue} \textbf{+}}1.581   & -0.045  & {\color{blue} \textbf{+}}0.088   \\ \hline
\multirow{9}{*}{\textit{YahooAnswer}} & LDA       & 0.190   & {\color{blue} \textbf{+}}0.030   & 0.278   & {\color{blue} \textbf{+}}0.086   & 0.378   & {\color{blue} \textbf{+}}0.018   & 0.510   & {\color{blue} \textbf{+}}0.266   & 0.063   & {\color{blue} \textbf{+}}0.023   \\
                                      & BTM       & 0.172   & {\color{blue} \textbf{+}}0.032   & 0.185   & {\color{blue} \textbf{+}}0.123   & 0.365   & {\color{blue} \textbf{+}}0.024   & 0.219   & {\color{blue} \textbf{+}}0.308   & 0.036   & {\color{blue} \textbf{+}}0.029   \\
                                      & G-BAT     & 0.213   & {\color{blue} \textbf{+}}0.022   & 0.265   & {\color{blue} \textbf{+}}0.121   & 0.399   & {\color{blue} \textbf{+}}0.016   & 0.304   & {\color{blue} \textbf{+}}0.426   & 0.062   & {\color{blue} \textbf{+}}0.028   \\
                                      & CNTM      & 0.294   & {\color{blue} \textbf{+}}0.007   & 0.534   & {\color{blue} \textbf{+}}0.020   & 0.455   & {\color{blue} \textbf{+}}0.006   & 1.311   & {\color{blue} \textbf{+}}0.109   & 0.141   & {\color{blue} \textbf{+}}0.006   \\
                                      & TSCTM     & 0.215   & {\color{blue} \textbf{+}}0.024   & 0.234   & {\color{blue} \textbf{+}}0.160   & 0.437   & {\color{blue} \textbf{+}}0.005   & -0.919  & {\color{blue} \textbf{+}}1.092   & 0.016   & {\color{blue} \textbf{+}}0.056   \\
                                      & BERTopic  & 0.252   & {\color{blue} \textbf{+}}0.006   & 0.421   & {\color{blue} \textbf{+}}0.036   & 0.448   & {\color{blue} \textbf{+}}0.003   & 0.533   & {\color{blue} \textbf{+}}0.197   & 0.094   & {\color{blue} \textbf{+}}0.011   \\
                                      & ECRTM     & 0.247   & {\color{blue} \textbf{+}}0.045   & 0.315   & {\color{blue} \textbf{+}}0.181   & 0.454   & {\color{blue} \textbf{+}}0.016   & -0.829  & {\color{blue} \textbf{+}}1.419   & 0.027   & {\color{blue} \textbf{+}}0.072   \\
                                      & CWTM      & 0.170   & {\color{blue} \textbf{+}}0.024   & 0.220   & {\color{blue} \textbf{+}}0.098   & 0.375   & {\color{blue} \textbf{+}}0.009   & 0.014   & {\color{blue} \textbf{+}}0.504   & 0.032   & {\color{blue} \textbf{+}}0.030   \\
                                      & LLM-TM    & 0.152   & {\color{blue} \textbf{+}}0.042   & 0.044   & {\color{blue} \textbf{+}}0.216   & 0.381   & {\color{blue} \textbf{+}}0.012   & -1.251  & {\color{blue} \textbf{+}}1.199   & -0.031  & {\color{blue} \textbf{+}}0.067   \\ \hline
\end{tabular}
}
\end{table*}

\subsection{Topic Quality Evaluation}
\paragraph{Topic Coherence.} We utilize base topic models with $K=20$ and $50$ for experiments in this part.
The average results, as depicted in Table~\ref{tab:main}, reveal the effectiveness of the Topic Refinement mechanism across various datasets and base models.
Notably, incorporating Topic Refinement consistently enhances the topic coherence of all base models, with increases in nearly every evaluated metric.
This improvement is especially evident in Tweet, AGNews, and TagMyNews datasets, where base models initially showed lower performance metrics.
In effect, Topic Refinement indirectly mitigates the challenges of data sparsity in short texts by refining topics.
While the LLM-TM approach offers potential, its effectiveness in generating topics in a bag-of-words format via prompts remains challenging, particularly for short texts. 
Even when the correct format is achieved, maintaining topic coherence can be problematic, underlining the necessity for further refinement to ensure quality.
Topic Refinement also boosts the performance of base models that initially exhibited high coherence, like CNTM and BERTopic. 

\paragraph{Topic Granularity.} Table~\ref{tab:embeddings} presents the results of within-topic similarity and between-topic distance.
The average improvements in $\mathcal{S}$ and $\mathcal{D}$ across all models and datasets are 21.23\% and 29.05\%, respectively.
Base models with higher topic granularity typically exhibit higher values of $\mathcal{S}$ and $\mathcal{D}$, indicating that their topics are more distinguishable and easier to understand. 
Models with lower initial topic granularity see greater increases from Topic Refinement.
Collectively, the advancements above prove the model-agnostic nature of our mechanism. 

\begin{table}[t]
\centering
\caption{Topic granularity results on four datasets.}
\label{tab:embeddings}
\resizebox{\linewidth}{!}{
\begin{tabular}{|c|cccc|cccc|cccc|cccc|}
    \hline
    \multirow{2}{*}{\textbf{\begin{tabular}[c]{@{}c@{}}Base\\ Models\end{tabular}}} & \multicolumn{4}{c|}{\textit{Tweet}}                 & \multicolumn{4}{c|}{\textit{AGNews}}                & \multicolumn{4}{c|}{\textit{TagMyNews}}             & \multicolumn{4}{c|}{\textit{YahooAnswer}}           \\ \cline{2-17} 
             & $\boldsymbol{\mathcal{S}}$ & \multicolumn{1}{c|}{$\boldsymbol{\Delta_{\mathcal{S}}}$} & $\boldsymbol{\mathcal{D}}$  & $\boldsymbol{\Delta_{\mathcal{D}}}$ & $\boldsymbol{\mathcal{S}}$ & \multicolumn{1}{c|}{$\boldsymbol{\Delta_{\mathcal{S}}}$} & $\boldsymbol{\mathcal{D}}$  & $\boldsymbol{\Delta_{\mathcal{D}}}$  & $\boldsymbol{\mathcal{S}}$ & \multicolumn{1}{c|}{$\boldsymbol{\Delta_{\mathcal{S}}}$} & $\boldsymbol{\mathcal{D}}$  & $\boldsymbol{\Delta_{\mathcal{D}}}$ & $\boldsymbol{\mathcal{S}}$ & \multicolumn{1}{c|}{$\boldsymbol{\Delta_{\mathcal{S}}}$} & $\boldsymbol{\mathcal{D}}$  & $\boldsymbol{\Delta_{\mathcal{D}}}$   \\ \hline
    LDA      & 0.254 & \multicolumn{1}{c|}{{\color{blue} \textbf{+}}0.048} & 13.12 & {\color{blue} \textbf{+}}4.02 & 0.279 & \multicolumn{1}{c|}{{\color{blue} \textbf{+}}0.045} & 12.25 & {\color{blue} \textbf{+}}4.13 & 0.280 & \multicolumn{1}{c|}{{\color{blue} \textbf{+}}0.053} & 14.17 & {\color{blue} \textbf{+}}4.47 & 0.368 & \multicolumn{1}{c|}{{\color{blue} \textbf{+}}0.034} & 15.93 & {\color{blue} \textbf{+}}4.01 \\
    BTM      & 0.259 & \multicolumn{1}{c|}{{\color{blue} \textbf{+}}0.048} & 14.03 & {\color{blue} \textbf{+}}3.86 & 0.267 & \multicolumn{1}{c|}{{\color{blue} \textbf{+}}0.067} & 10.41 & {\color{blue} \textbf{+}}6.13 & 0.249 & \multicolumn{1}{c|}{{\color{blue} \textbf{+}}0.083} & 11.73 & {\color{blue} \textbf{+}}6.64 & 0.345 & \multicolumn{1}{c|}{{\color{blue} \textbf{+}}0.037} & 11.30 & {\color{blue} \textbf{+}}4.96 \\
    G-BAT    & 0.248 & \multicolumn{1}{c|}{{\color{blue} \textbf{+}}0.058} & 14.21 & {\color{blue} \textbf{+}}4.06 & 0.288 & \multicolumn{1}{c|}{{\color{blue} \textbf{+}}0.056} & 15.44 & {\color{blue} \textbf{+}}3.79 & 0.328 & \multicolumn{1}{c|}{{\color{blue} \textbf{+}}0.046} & 19.29 & {\color{blue} \textbf{+}}3.43 & 0.367 & \multicolumn{1}{c|}{{\color{blue} \textbf{+}}0.039} & 18.77 & {\color{blue} \textbf{+}}3.24 \\
    CNTM     & 0.253 & \multicolumn{1}{c|}{{\color{blue} \textbf{+}}0.046} & 14.35 & {\color{blue} \textbf{+}}3.39 & 0.327 & \multicolumn{1}{c|}{{\color{blue} \textbf{+}}0.038} & 22.31 & {\color{blue} \textbf{+}}2.95 & 0.341 & \multicolumn{1}{c|}{{\color{blue} \textbf{+}}0.035} & 25.88 & {\color{blue} \textbf{+}}2.59 & 0.446 & \multicolumn{1}{c|}{{\color{blue} \textbf{+}}0.008} & 27.97 & {\color{blue} \textbf{+}}0.83 \\
    BERTopic & 0.274 & \multicolumn{1}{c|}{{\color{blue} \textbf{+}}0.033} & 16.18 & {\color{blue} \textbf{+}}2.57 & 0.237 & \multicolumn{1}{c|}{{\color{blue} \textbf{+}}0.090} & 16.45 & {\color{blue} \textbf{+}}6.17 & 0.255 & \multicolumn{1}{c|}{{\color{blue} \textbf{+}}0.080} & 20.62 & {\color{blue} \textbf{+}}5.41 & 0.318 & \multicolumn{1}{c|}{{\color{blue} \textbf{+}}0.056} & 21.42 & {\color{blue} \textbf{+}}3.87 \\
    TSCTM    & 0.252 & \multicolumn{1}{c|}{{\color{blue} \textbf{+}}0.061} & 14.46 & {\color{blue} \textbf{+}}4.81 & 0.288 & \multicolumn{1}{c|}{{\color{blue} \textbf{+}}0.048} & 19.42 & {\color{blue} \textbf{+}}3.59 & 0.279 & \multicolumn{1}{c|}{{\color{blue} \textbf{+}}0.041} & 19.78 & {\color{blue} \textbf{+}}4.14 & 0.381 & \multicolumn{1}{c|}{{\color{blue} \textbf{+}}0.012} & 25.22 & {\color{blue} \textbf{+}}0.99 \\
    ECRTM    & 0.245 & \multicolumn{1}{c|}{{\color{blue} \textbf{+}}0.060} & 16.46 & {\color{blue} \textbf{+}}4.33 & 0.218 & \multicolumn{1}{c|}{{\color{blue} \textbf{+}}0.122} & 15.18 & {\color{blue} \textbf{+}}7.94 & 0.243 & \multicolumn{1}{c|}{{\color{blue} \textbf{+}}0.113} & 18.24 & {\color{blue} \textbf{+}}7.80 & 0.333 & \multicolumn{1}{c|}{{\color{blue} \textbf{+}}0.065} & 23.46 & {\color{blue} \textbf{+}}4.14 \\
    CWTM     & 0.273 & \multicolumn{1}{c|}{{\color{blue} \textbf{+}}0.062} & 15.32 & {\color{blue} \textbf{+}}5.80 & 0.296 & \multicolumn{1}{c|}{{\color{blue} \textbf{+}}0.033} & 14.50 & {\color{blue} \textbf{+}}2.40 & 0.304 & \multicolumn{1}{c|}{{\color{blue} \textbf{+}}0.032} & 16.53 & {\color{blue} \textbf{+}}2.25 & 0.347 & \multicolumn{1}{c|}{{\color{blue} \textbf{+}}0.028} & 15.26 & {\color{blue} \textbf{+}}2.57 \\
    LLM-TM   & 0.218 & \multicolumn{1}{c|}{{\color{blue} \textbf{+}}0.076} & 11.17 & {\color{blue} \textbf{+}}5.06 & 0.231 & \multicolumn{1}{c|}{{\color{blue} \textbf{+}}0.091} & 11.68 & {\color{blue} \textbf{+}}5.81 & 0.234 & \multicolumn{1}{c|}{{\color{blue} \textbf{+}}0.108} & 11.89 & {\color{blue} \textbf{+}}7.12 & 0.263 & \multicolumn{1}{c|}{{\color{blue} \textbf{+}}0.071} & 12.31 & {\color{blue} \textbf{+}}5.65 \\ \hline
\end{tabular}
}
\end{table}

\begin{table}[t]
    \centering
    \caption{Average token costs for each topic.}
    \label{tab:cost}
    \resizebox{0.5\linewidth}{!}{
\begin{tabular}{|c|cc|cc|}
\hline
\multirow{2}{*}{\textbf{Datasets}} & \multicolumn{2}{c|}{\textbf{Ours}}                 & \multicolumn{2}{c|}{\textbf{LLM-TM}}                                    \\
                                   & \textbf{\#Input} & \textbf{\#Output} & \textbf{\#Input} & \textbf{\#Output} \\ \hline
\textit{Tweet}                     & 1591.01                 & 274.90                   & 4958.73                 & 1214.07                                      \\
\textit{AGNews}                    & 1588.60                 & 264.22                   & 33382.61                & 7541.41                                      \\
\textit{TagMyNews}                 & 1593.70                 & 275.83                   & 61872.93                & 14448.79                                     \\
\textit{YahooAnswer}               & 1578.43                 & 231.50                   & 27358.56                & 5913.36                                      \\ \hline
\end{tabular}}
\end{table}
\begin{table}[t]
\centering
\caption{Performance results of text classification on four datasets. \textit{Rate}$\uparrow$ means the growth rate of Topic Refinement.}
\label{tab:class}
\resizebox{\linewidth}{!}{
\begin{tabular}{|c|cccc|cccc|cccc|cccc|}
    \hline
    \multirow{2}{*}{\textbf{\begin{tabular}[c]{@{}c@{}}Base\\ Models\end{tabular}}} & \multicolumn{4}{c|}{\textit{Tweet}}                             & \multicolumn{4}{c|}{\textit{AGNews}}                            & \multicolumn{4}{c|}{\textit{TagMyNews}}                         & \multicolumn{4}{c|}{\textit{YahooAnswer}}                     \\ \cline{2-17} 
              & \textbf{Acc} & \multicolumn{1}{c|}{\textbf{\textit{Rate}} $\boldsymbol{\uparrow}$}  & \textbf{F1} & \textbf{\textit{Rate}} $\boldsymbol{\uparrow}$  & \textbf{Acc} & \multicolumn{1}{c|}{\textbf{\textit{Rate}} $\boldsymbol{\uparrow}$}  & \textbf{F1} & \textbf{\textit{Rate}} $\boldsymbol{\uparrow}$  & \textbf{Acc} & \multicolumn{1}{c|}{\textbf{\textit{Rate}} $\boldsymbol{\uparrow}$}  & \textbf{F1} & \textbf{\textit{Rate}} $\boldsymbol{\uparrow}$  & \textbf{Acc} & \multicolumn{1}{c|}{\textbf{\textit{Rate}} $\boldsymbol{\uparrow}$} & \textbf{F1} & \textbf{\textit{Rate}} $\boldsymbol{\uparrow}$ \\ \hline
    LDA       & 0.913        & \multicolumn{1}{c|}{16.98\%}     & 0.720       & 18.19\%     & 0.705        & \multicolumn{1}{c|}{12.34\%}    & 0.702       & 10.26\%     & 0.705        & \multicolumn{1}{c|}{9.22\%}      & 0.676       & 9.76\%     & 0.621     & \multicolumn{1}{c|}{0.64\%}    & 0.614       & 1.14\% \\
    BTM       & 0.897        & \multicolumn{1}{c|}{17.50\%}     & 0.692       & 9.54\%      & 0.701        & \multicolumn{1}{c|}{5.56\%}     & 0.706       & 5.95\%      & 0.686        & \multicolumn{1}{c|}{11.81\%}     & 0.656       & 12.35\%    & 0.616     & \multicolumn{1}{c|}{0.32\%}    & 0.610       & 0.66\% \\
    G-BAT     & 0.874        & \multicolumn{1}{c|}{19.57\%}     & 0.672       & 26.79\%     & 0.698        & \multicolumn{1}{c|}{3.87\%}     & 0.703       & 3.98\%      & 0.680        & \multicolumn{1}{c|}{1.03\%}      & 0.645       & 0.93\%     & 0.610     & \multicolumn{1}{c|}{1.15\%}    & 0.602       & 1.50\% \\
    CNTM      & 0.881        & \multicolumn{1}{c|}{16.00\%}     & 0.629       & 10.02\%     & 0.726        & \multicolumn{1}{c|}{3.31\%}     & 0.730       & 3.84\%      & 0.701        & \multicolumn{1}{c|}{11.70\%}     & 0.672       & 12.65\%    & 0.628     & \multicolumn{1}{c|}{3.34\%}    & 0.621       & 3.86\% \\
    BERTopic  & 0.897        & \multicolumn{1}{c|}{23.19\%}     & 0.683       & 16.54\%     & 0.704        & \multicolumn{1}{c|}{1.70\%}     & 0.707       & 1.70\%      & 0.690        & \multicolumn{1}{c|}{9.57\%}      & 0.655       & 9.92\%     & 0.619     & \multicolumn{1}{c|}{1.62\%}    & 0.612       & 1.96\% \\
    TSCTM     & 0.895        & \multicolumn{1}{c|}{15.20\%}     & 0.688       & 9.88\%      & 0.718        & \multicolumn{1}{c|}{5.85\%}     & 0.715       & 5.03\%      & 0.693        & \multicolumn{1}{c|}{7.79\%}      & 0.660       & 7.88\%     & 0.626     & \multicolumn{1}{c|}{0.96\%}    & 0.618       & 0.81\% \\
    ECRTM     & 0.897        & \multicolumn{1}{c|}{9.70\%}      & 0.675       & 8.30\%      & 0.587        & \multicolumn{1}{c|}{3.41\%}     & 0.593       & 5.06\%      & 0.650        & \multicolumn{1}{c|}{6.77\%}      & 0.617       & 8.10\%     & 0.598     & \multicolumn{1}{c|}{3.34\%}    & 0.592       & 4.90\% \\
    CWTM      & 0.848        & \multicolumn{1}{c|}{22.88\%}     & 0.568       & 21.13\%     & 0.726        & \multicolumn{1}{c|}{10.61\%}    & 0.727       & 10.45\%     & 0.683        & \multicolumn{1}{c|}{16.54\%}     & 0.651       & 15.21\%    & 0.617     & \multicolumn{1}{c|}{3.57\%}    & 0.611       & 3.93\% \\
    LLM-TM    & 0.883        & \multicolumn{1}{c|}{11.10\%}     & 0.658       & 8.66\%      & 0.729        & \multicolumn{1}{c|}{6.45\%}     & 0.730       & 5.89\%      & 0.686        & \multicolumn{1}{c|}{4.23\%}      & 0.650       & 3.69\%     & 0.619     & \multicolumn{1}{c|}{1.45\%}    & 0.610       & 1.64\% \\ \hline
    \end{tabular}
}

\end{table}

\paragraph{Token Cost.}
Table~\ref{tab:cost} presents the average token costs for each topic across various datasets, showcasing the efficiency of our mechanism.
The input token costs are relatively consistent across datasets due to the similar prompts.
This uniformity contrasts with existing LLM-based topic modeling methods, where token costs often scale with dataset size. 
Our approach, however, links token costs primarily to the number of topics rather than dataset size, which is more economical.
Interestingly, the output token costs suggest differences in the complexity of topic modeling for each dataset. 
For instance, the YahooAnswer dataset, which already exhibited relatively higher-quality topics in the base models, requires fewer tokens for refinement.
The implication is that datasets with lower modeling complexity require less intervention from our mechanism.

\subsection{Text Classification}
We also evaluate the performance improvement in text classification tasks achieved by refining base topics. 
Our experiments utilize the SentenceTransformer\footnote{\url{https://huggingface.co/sentence-transformers/all-MiniLM-L12-v2}} model to generate embeddings for each document and the topics, and compute the cosine similarity between the document embedding and topic embeddings. 
The normalized cosine similarities are taken as the topic distribution of this document.
The topic distributions in the dataset are divided into an 8:2 training-testing split and employ an SVM classifier for text classification.
The results, outlined in Table~\ref{tab:class}, confirm that Topic Refinement can enhance the performance of topic-related text classification tasks.
We can see that base models that effectively capture high-quality topics generally exhibit better classification performance. 
This trend is more pronounced in larger datasets. 
Moreover, datasets with notable improvements in topic quality, such as Tweet, show a more significant increase in classification outcomes post-refinement. 
These findings suggest a robust capability for refining the utility of topics, which benefits the ability of the SVM classifier to discern relevant textual features.

\begin{figure*}[!t]
  \centering
  \includegraphics[width=0.8\linewidth]{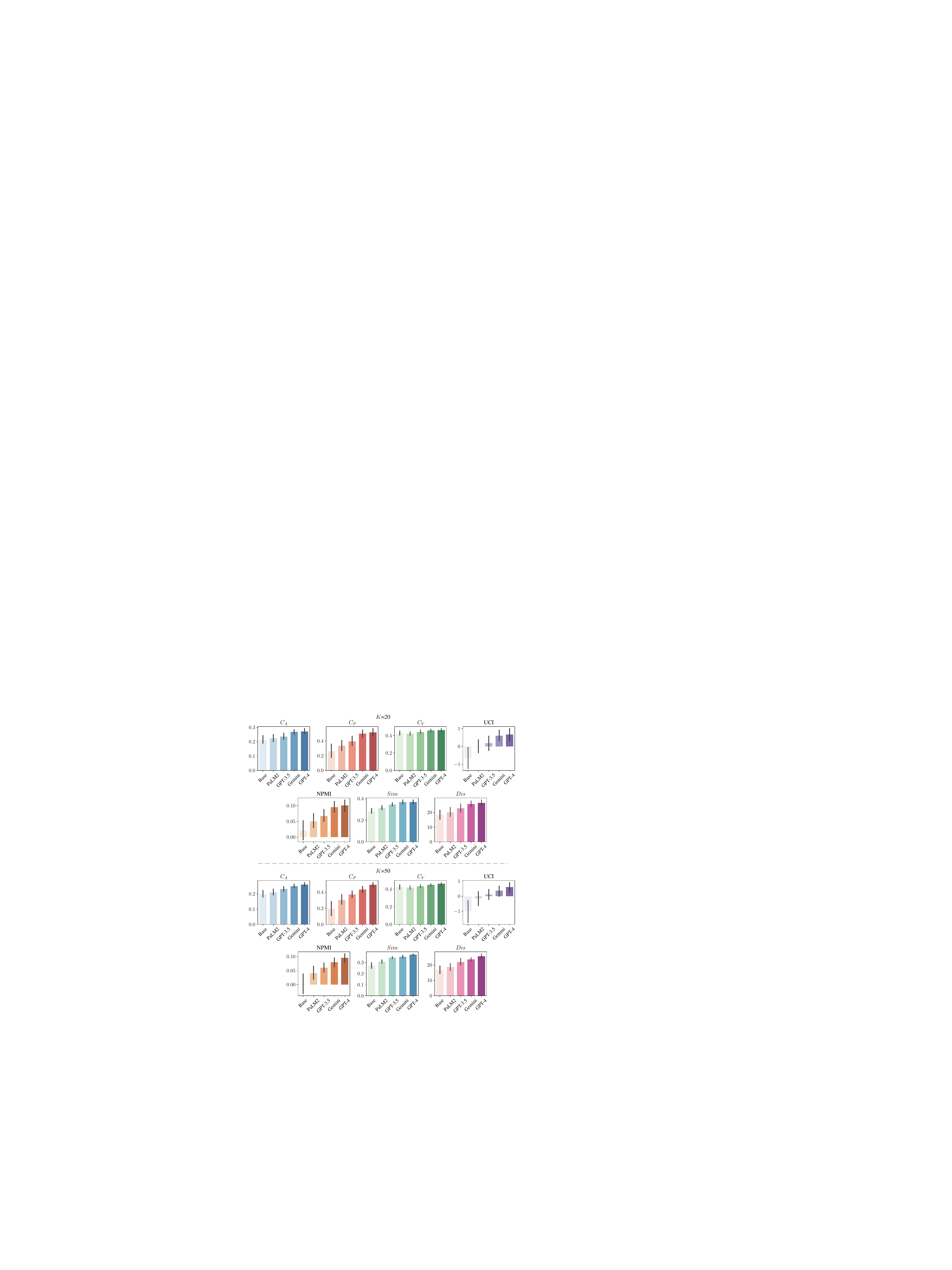}
  \caption{Ablation study results with various LLMs. The error bars denote a 95\% confidence interval for the statistical variability of results across nine base models.}
  \label{fig:llms}
\end{figure*}

\begin{figure}[t]
    \centering
    \includegraphics[width=0.7\linewidth]{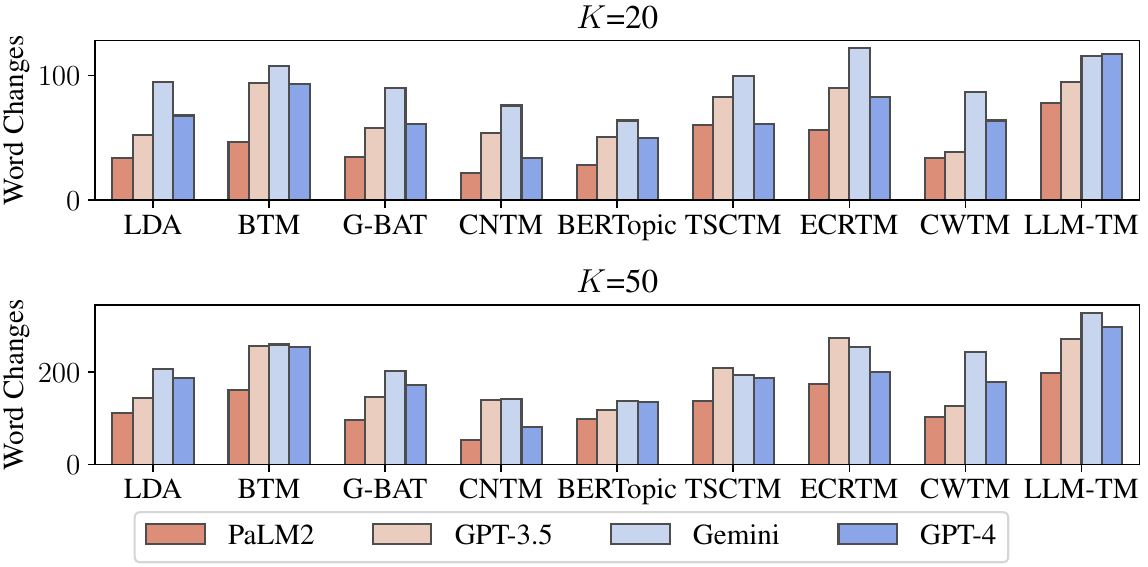}
    \caption{Number of word changes between base topics and refined topics.}
    \label{fig:wordchanges}
\end{figure}

\begin{figure}[ht]
    \centering
    \includegraphics[width=0.65\linewidth]{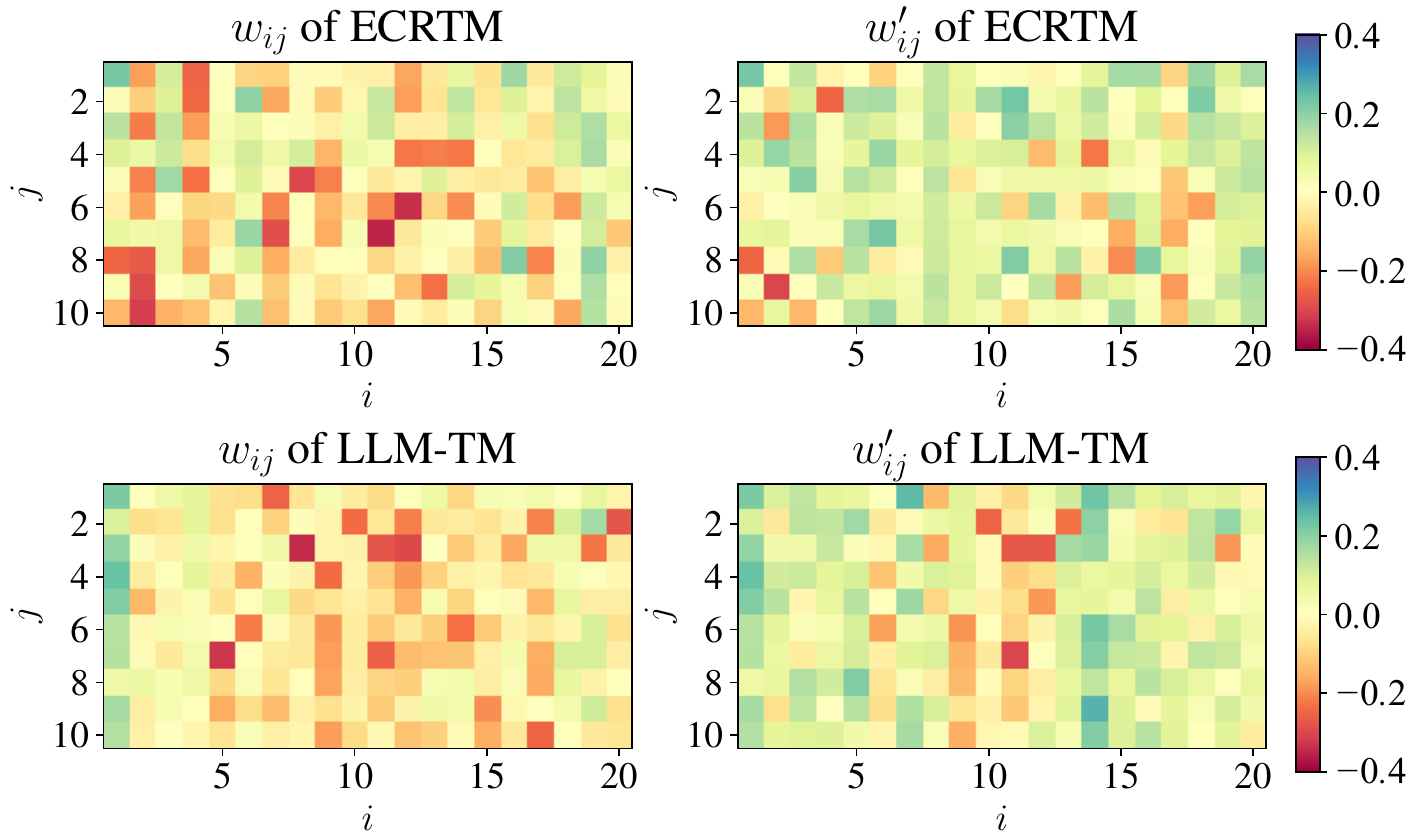}
    \caption{Visualization of NPMI for word $w_{ij}$ and $w'_{ij}$.}
    \label{fig:heatmap}
\end{figure}

% \begin{figure}[ht]
%     \centering
%     \includegraphics[width=\linewidth]{NPMI_results_scatter_combined.pdf}
%     \caption{Visualization of average NPMI for topic $t_i$ and $t'_i$.}
%     \vspace{-0.15cm}
%     \label{fig:scatter}
% \end{figure}

\begin{table}[t]
\centering
\caption{Case study of base and refined topics. The replaced words in base topics are in {\color{red}red}, and the alternative words in refined topics are in {\color{blue}blue}.}
\label{tab:case}
\resizebox{0.8\linewidth}{!}{
\begin{tabular}{|l|}
     \hline
     \textbf{Topic 1: Finance}                                                                \\ \hline
     wealth billion fund private repay yuan \textit{\color{red} lcd} \textit{\color{red} mutual} shareholder refund        \\
     wealth billion fund private repay yuan \textit{\color{blue} budget} \textit{\color{blue} investment} shareholder refund \\ \hline
     \textbf{Topic 2: Theater}                                                                \\ \hline
     review theater \textit{\color{red} elizabeth} \textit{\color{red} taylor} shakespeare \textit{\color{red} shore} \textit{\color{red} leo} \textit{\color{red} love} \textit{\color{red} gil} \textit{\color{red} dragon}       \\
     review theater \textit{\color{blue} performance} \textit{\color{blue} director} shakespeare \textit{\color{blue} play} \textit{\color{blue} actor} \textit{\color{blue} drama} \textit{\color{blue} act} \textit{\color{blue} stage}  \\ \hline
     \textbf{Topic 3: Investment}                                                               \\ \hline
     \textit{\color{red} georgia} investor return \textit{\color{red} legacy} bank \textit{\color{red} society} \textit{\color{red} android} \textit{\color{red} editorial} \textit{\color{red} math} \textit{\color{red} critic} \\ 
     \textit{\color{blue} credit} investor return \textit{\color{blue} wealth} bank \textit{\color{blue} economy} \textit{\color{blue} stock} \textit{\color{blue} investment} \textit{\color{blue} money} \textit{\color{blue} budget} \\ \hline
     \textbf{Topic 4: Medicine}                                                                \\ \hline
     treatment spread relief stress eye \textit{\color{red} pollution} \textit{\color{red} power} wound \textit{\color{red} shark} \textit{\color{red} tournament} \\ 
     treatment spread relief stress eye \textit{\color{blue} infection} \textit{\color{blue} illness} wound \textit{\color{blue} doctor} \textit{\color{blue} medicine} \\ \hline
\end{tabular}
}
\end{table}

\subsection{Ablation Study}
We additionally carry out ablation studies to evaluate the impact of different LLMs, such as PaLM2, GPT-3.5, Gemini Pro, and GPT-4, on the TagMyNews dataset with $K=20 \text{ and } 50$.
Figure~\ref{fig:llms} compares quality metrics and the number of altered words within the topics after Topic Refinement.
All tested LLMs clearly improve topic quality, underscoring the efficacy and robustness of our mechanism.
Notably, LLMs with more recent and superior capabilities exhibit better refinement performance, with GPT-4 emerging as the most effective among them.
Figure~\ref{fig:wordchanges} illustrates the number of word changes in $K$ topics after Topic Refinement.
Although Gemini Pro and GPT-4 yield comparable results in quality metrics, GPT-4 shows fewer topic alterations, indicating its precision in refining topics.

\subsection{Visualization and Case Study}
To provide an intuitive understanding of how Topic Refinement affects topic quality, we display the coherence visualization of the topics mined by ECRTM and LLM-TM on the TagMyNews dataset with $K=20$.
These two base models are selected because they exhibit the most word changes upon refinement.
In the visualization, we compute the metric NPMI for each word in a topic against the other $N - 1$ words to assess word coherence.
Figure~\ref{fig:heatmap} compares the NPMI for individual words within the topics before and after refinement, indicated by $w_{ij}$ and $w'_{ij}$, respectively.
The heatmaps show a visible shift towards higher values after refinement to prove the enhancement in word-topic relatedness.
% Figure~\ref{fig:scatter} further contrasts the average NPMI from base topics to refined topics, while the refined topics generally reveal higher NPMI values. 
% The error bars with a 95\% confidence interval prove this improvement and are tighter for the refined topics. 
Further, we list some base and refined topics mined from ECRTM (Topic 1 and 2) and LLM-TM (Topic 3 and 4) to display the results before and after refinement directly.
From Table~\ref{tab:case}, we can see that the Topic Refinement mechanism directly improves the quality of the topics.

\section{Conclusion}
\label{sec:conclusion}
This paper explored the potential of using LLMs to enhance the topic modeling quality in short texts.
Our proposed mechanism, Topic Refinement, utilizes prompt engineering with LLMs to identify and correct semantically intruder words within extracted topics. 
This process mimics human-like evaluation and refinement of topics to enhance the topic quality.
Extensive experiments across the four datasets and eight base topic models conclusively demonstrated the effectiveness of this mechanism. 
As the initial topic quality of base models influences the final performance, future research could involve the dynamic adjustment of the topic modeling process based on LLM feedback.

\begin{credits}
\subsubsection{\ackname} This work was partially supported by the Major Research Plan of the National Natural Science Foundation of China under Grant No. 92467202, the National Natural Science Foundation of China under Grant No. 62102192, the Fellowship of China Postdoctoral Science Foundation under Grant No. 2022M710071, the Open Fund of Anhui Province Key Laboratory of Cyberspace Security Situation Awareness and Evaluation under Grant No. TK224013, and the Postgraduate Research and Practice Innovation Program of Jiangsu Province under Grant No. KYCX23\_1077.

\end{credits}

% ---- Bibliography ----
%
% BibTeX users should specify bibliography style 'splncs04'.
% References will then be sorted and formatted in the correct style.
%
\bibliographystyle{splncs04}
\bibliography{reference}
\end{document}